%% file: egpaper_final.tex
\documentclass[10pt,twocolumn,letterpaper]{article}
\pdfoutput=1
\usepackage{iccv}
\usepackage{times}
\usepackage{epsfig}
\usepackage{graphicx}
\usepackage{amsmath}
\usepackage{amssymb}
\usepackage{soul}
\usepackage{overpic}
\usepackage{xcolor}
\usepackage{tabu}
\usepackage{placeins}
\usepackage{booktabs}
\usepackage{setspace}
\usepackage{enumitem}
\usepackage{bbm}
\usepackage{boldline}
\usepackage{array,multirow}
\usepackage{soul}
\usepackage{pifont}
\usepackage[export]{adjustbox}
\usepackage[accsupp]{axessibility}  
\input{defines}

\newcommand{\cmark}{\ding{51}}%
\newcommand{\xmark}{\ding{55}}%
\newcommand{\myparagraph}[1]{\vspace{2pt}\noindent{\textbf{#1}}}
\newcolumntype{x}[1]{>{\centering\arraybackslash\hspace{0pt}}p{#1}}

\def\ourda{ViSGA\xspace}

\usepackage[pagebackref=true,breaklinks=true,letterpaper=true,colorlinks,bookmarks=false]{hyperref}

\iccvfinalcopy 

\def\iccvPaperID{8560} 

\ificcvfinal\fi

\begin{document}

\title{Seeking Similarities over Differences: Similarity-based Domain Alignment for Adaptive Object Detection}

\author{
	Farzaneh Rezaeianaran\textsuperscript{1} \ \ \ 
	Rakshith Shetty\textsuperscript{1} \ \ \ 
	Rahaf Aljundi\textsuperscript{2} \\
	Daniel Olmeda Reino\textsuperscript{2}\ \ \ 
	Shanshan Zhang\textsuperscript{3} \ \ \ 
	Bernt Schiele\textsuperscript{1} \\
	\small \textsuperscript{1}Max Planck Institute for Informatics, Saarland Informatics Campus \ \ \ \ \ \textsuperscript{2}Toyota Motor Europe \ \ \ \ \
	\textsuperscript{3}Nanjing University of Science and Technology
}



\maketitle
\ificcvfinal\thispagestyle{empty}\fi
\begin{abstract}
In order to robustly deploy object detectors across a wide range of scenarios, they should be adaptable to shifts in the input distribution without the need to constantly annotate new data. 
This has motivated research in Unsupervised Domain Adaptation~(UDA) algorithms for detection.
UDA methods learn to adapt from labeled source domains to unlabeled target domains, by inducing alignment between detector features from source and target domains. 
Yet, there is no consensus on what features to align and how to do the alignment.
%
In our work, we propose a framework that generalizes the different components commonly used by UDA methods laying the ground for an in-depth analysis of the UDA design space.
%
Specifically, we propose a novel UDA algorithm, \ourda, a direct implementation of our framework, that leverages the best design choices and introduces a simple but effective method to aggregate features at instance-level based on visual similarity before inducing group alignment via adversarial training.
%
We show that both similarity-based grouping and adversarial training allows our model to focus on coarsely aligning feature groups, without being forced to match all instances across loosely aligned domains.
Finally, we examine the applicability of \ourda to the setting where labeled data are gathered from different sources.
%
Experiments show that not only our method outperforms previous single-source approaches on Sim2Real and Adverse Weather, but also generalizes well to the multi-source setting.
\end{abstract}

\section{Introduction}
\label{sec:introduction}
\input{sections/introduction}
\section{Related Work}
\label{sec:related}
\input{sections/related}

\section{Our General Framework for UDA}
\label{sec:components}
\input{sections/components}


\section{Experiments}
\label{sec:analysis}
\input{sections/analysis}


\section{Generalization to Multi-Source}
\label{sec:generalization}
\input{sections/generalization}

\section{Conclusions}
\label{sec:conclusion}
\input{sections/conclusions}


{\small
\bibliographystyle{ieee_fullname}
\bibliography{egbib}
}

\end{document}

%% file: defines.tex
 \makeatletter
 \DeclareRobustCommand\onedot{\futurelet\@let@token\@onedot}
 \def\@onedot{\ifx\@let@token.\else.\null\fi\xspace}

 \makeatother

\DeclareRobustCommand{\figref}[1]{Figure~\ref{#1}}

\DeclareRobustCommand{\Figref}[1]{Figure~\ref{#1}}

\DeclareRobustCommand{\secref}[1]{Section~\ref{#1}}

\DeclareRobustCommand{\Tableref}[1]{Table~\ref{#1}}

\DeclareRobustCommand{\eqnref}[1]{Equation~(\ref{#1})}

\DeclareRobustCommand{\eqnsref}[1]{Equations~(\ref{#1})}

%% file: sections/introduction.tex
\begin{figure}[t]
	\centering
	\includegraphics[width=0.90\linewidth]{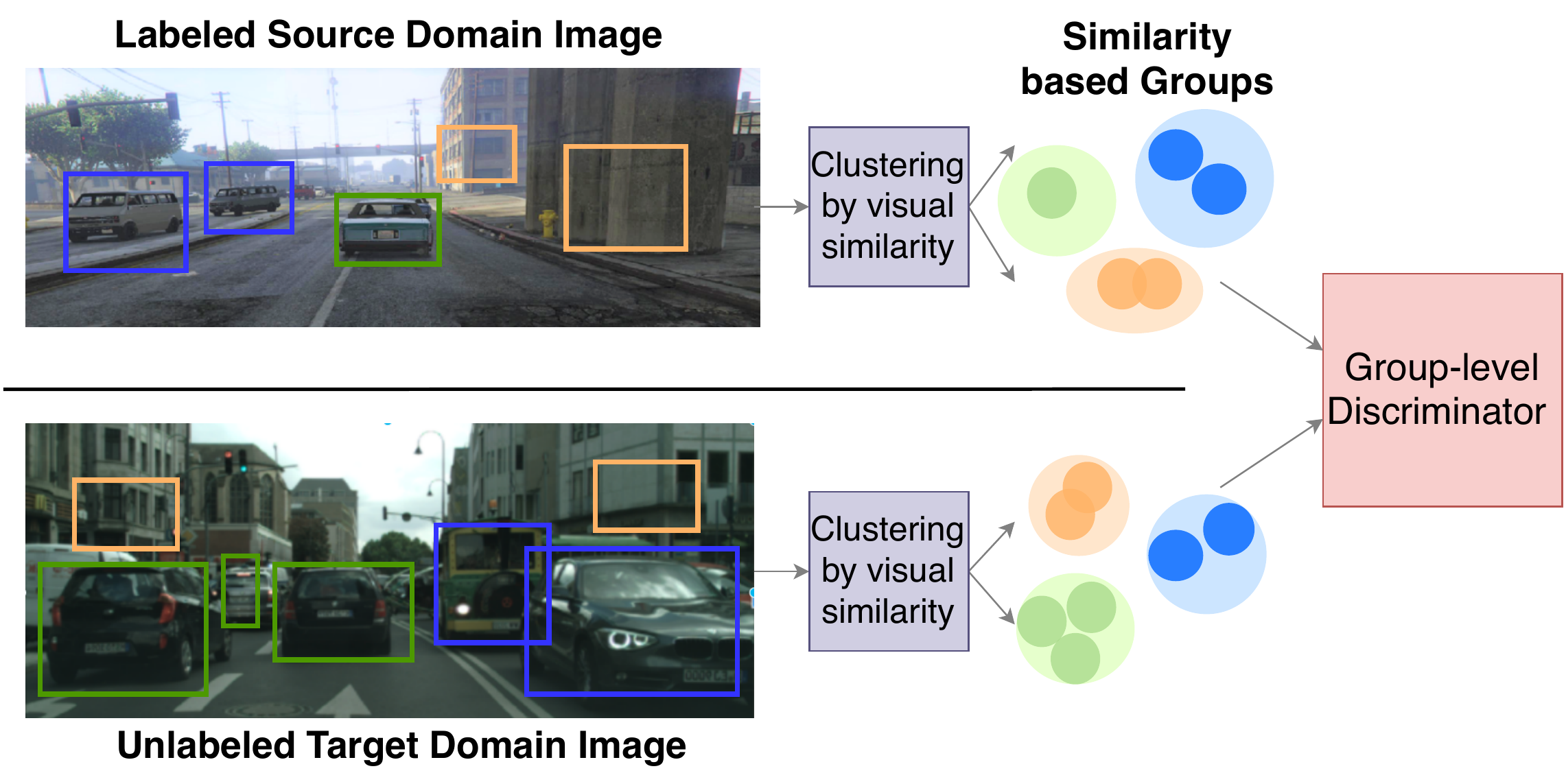}
    \caption{Depiction of visual similarity based grouping proposed in our \ourda method. Instance proposals from the detector are aggregated based on visual similarity to create an adaptive number of class-agnostic groups then they are aligned across the domains.
    } 
	\label{fig:teaser}
	\vspace{-1mm}
\end{figure}
Object detectors should be adaptable to ``domain shift'' that can occur due to many factors including changes in weather or camera, compared to the training data. 
Domain shifts can cause a significant drop in object detector performance~\cite{da_faster_rcnn, gopalan2011domain}.  
Domain adaptation methods~\cite{duan2011visual, duan2012domain, tzeng2015simultaneous, long2015learning, long2017deep, motiian2017unified}  study this problem, casting it as a task of learning models from a source domain and adapting to a target domain. 
In object detection, where collecting bounding box annotations is expensive, it becomes critical that domain adaptation can be performed without the need to annotate every new domain.
This motivates the challenging setting of unsupervised domain adaptation~(UDA)~\cite{survey_2018, tzeng2017adversarial, lu2017unsupervised, cariucci2017autodial}, where one has access to labeled source data and only unlabeled target data. 
Moreover, training data itself could be gathered under different conditions, a scenario typically referred to as a multi-source domain adaptation~\cite{multipeng2019moment, multixu2018deep, multizhao2018adversarial, multizhao2019}.

A dominant line in UDA works is to learn invariant representations via aligning source and target domains, with various proposed alignment strategies.
Specifically in object detection, the questions of what features to align and how to induce the alignment have been the subject of recent research. 
Early works~\cite{da_faster_rcnn, he_iccv19_MAF} propose aligning both image-level features from the backbone network and all instance-level features extracted from object proposals using adversarial training~\cite{grl_ganin}. 
A recent state-of-the-art approach~\cite{GPA} argues that it is beneficial to aggregate object proposals before alignment and suggests condensing all proposals into a single category prototype vector before inducing alignment using a contrastive loss.
This raises questions on what is the right aggregation-level at which to do feature alignment and what is the right mechanism to induce this alignment.
In this work, we propose a novel UDA method for object detection, called visually similar group alignment~(\ourda). Our method harnesses the power of adversarial training, while leveraging the visual similarity of the different proposals as a basis for aggregating them. 
By relying on visual similarity, we aggregate proposals from potentially different spatial locations~(Figure~\ref{fig:teaser}), increasing the effectiveness of adversarial training.
Doing so, we drive a more powerful discriminator and hence better aligned features. 
To enhance the flexibility of proposal aggregation and to avoid introducing unwanted noise in the alignment process as a result of a preset fixed number of groups, we opt for  dynamic clustering based on the distance at which proposals are aggregated. This
 improves the adaptability of our method to a variable number of objects present in the input.  

Our method design choices are based on an in-depth analysis of common components of UDA methods for detection.
%
In particular we study what is the right aggregation-level to perform instance-level alignment, ranging from considering all instances~\cite{da_faster_rcnn}, multiple groups based on clustering to single prototypes~\cite{GPA}. 
When aggregating object proposals, we analyze whether including the predicted class label is beneficial and which distance metric performs better, including spatial overlap and visual similarity.
We further compare the effectiveness of using contrastive losses versus adversarial training, as the alignment mechanism. 
%
%
%
%

In summary, our key contributions are as follows:
1) We propose a novel, simple yet effective, UDA method for object detection via adversarial training and dynamic visual similarity-based grouping of proposals from the source and target domains.
2) We perform an in-depth analysis answering questions on what is the right level of alignment and how to induce alignment. 
3) We evaluate our proposed approach on three different domain shift scenarios including: Adverse weather, Synthetic to Real data, and Cross camera and show state-of-the-art results. 
4) We are the first to consider the important setting of multi-source domain adaptation for object detection where annotated data are gathered from different sources. We show that our method continues to improve  in this highly relevant scenario, another evidence for the effectiveness of our approach. 

%% file: sections/related.tex
\myparagraph{Object detection.} Classical object detection methods were based on sliding window classification using hand-crafted features~\cite{dalal2005histograms, viola2001rapid, felzenszwalb2009object}. 
However, deep convolutional networks (CNNs)~\cite{krizhevsky2017imagenet, he2016deep, simonyan2014very} trained on large scale data~\cite{Chen2015, pascal} have become popular recently. 
These can be categorized into one- \cite{liu2016ssd, redmon2016you,redmon2017yolo9000} and two-stage frameworks \cite{girshick2014rich, girshick2015fast, he2015spatial, ren2015faster}. Among them Faster R-CNN \cite{ren2015faster} is widely adopted due to good performance and good open implementations. 
Faster R-CNN extends prior works~\cite{girshick2014rich, girshick2015fast} with a Region Proposal Network (RPN). 
A second detection head classifies regions of interest (RoI) and is trained end-to-end with RPN. In our work, we use Faster R-CNN as our base detector. 

\myparagraph{Unsupervised domain adaptation for object detection.} Chen et al.\ \cite{da_faster_rcnn} is an early UDA method for object detection. It proposes to learn domain-invariant features at both image and instance-level using adversarial training (AT)~\cite{grl_ganin} on top of the Faster R-CNN detector.
%
This idea motivates other works, that focus on selecting the right features and right level of aggregation for alignment \cite{strong-weak, he_iccv19_MAF, zhu_cvpr19_selective_alignment, xu_cvpr20_icr_ccr, chen_cvpr20_htcn}. Both \cite{strong-weak, he_iccv19_MAF} adapt adversarial strategy to align image-level features.
while, He et al.\ \cite{he_iccv19_MAF}, employ multiple domain discriminators and they also encode class information together with features for the instance level alignment. 
Xu et al.\ \cite{xu_cvpr20_icr_ccr} add a categorical classifier for image-level alignment to weakly learn class features with source domain supervision. On the other hand, some recent works have proposed applying different alignment mechanisms \cite{ zhuang_2020_ifan, zheng_cvpr20_prototype, GPA}. Xu et al.\ \cite{GPA} employ a geometry-based prototype construction and use contrastive losses instead of AT for learning domain invariant features. 
Similar contrastive losses were applied in training domain adaptive classifiers in \cite{kang2019contrastive}.
Zheng et al.\ \cite{zheng_cvpr20_prototype}, propose a hybrid framework to minimize $L2$ distance between single-class specific prototypes across domains at instance-level and using adversarial training at image-level.
%

In this paper, we propose a novel framework \ourda by leveraging the best design practices from prior work. Unlike \cite{GPA, zheng_cvpr20_prototype}, our approach uses a similarity-based grouping scheme to aggregate information into multiple groups in a class agnostic manner. In addition, we purely use an adversarial strategy unlike a hybrid framework used by \cite{zheng_cvpr20_prototype} or Contrastive losses used by \cite{GPA}. 

Moreover, to the best of our knowledge, existing UDA methods for detection, only consider the single-source UDA. Recently, a line of work using deep models is proposed for multi-source setting, where the training data are collected from multiple sources~\cite{multipeng2019moment, multixu2018deep, multizhao2018adversarial, multizhao2019}. These works mainly consider image classification, except~\cite{multipeng2019moment} which is proposed for semantic segmentation. The general idea of these works is to consider additional components or computations to align each source domain to the target~\cite{ multixu2018deep, multizhao2018adversarial, multizhao2019} or aggregate information from all of the sources into one before adapting to the target domain~\cite{multipeng2019moment}. In this work, besides single-source UDA, we consider the generalization of our method to multi-source to further examine the effectiveness of our general framework.

%% file: sections/components.tex
\begin{figure*}[t]
	\centering
	\includegraphics[width=0.65 \linewidth]{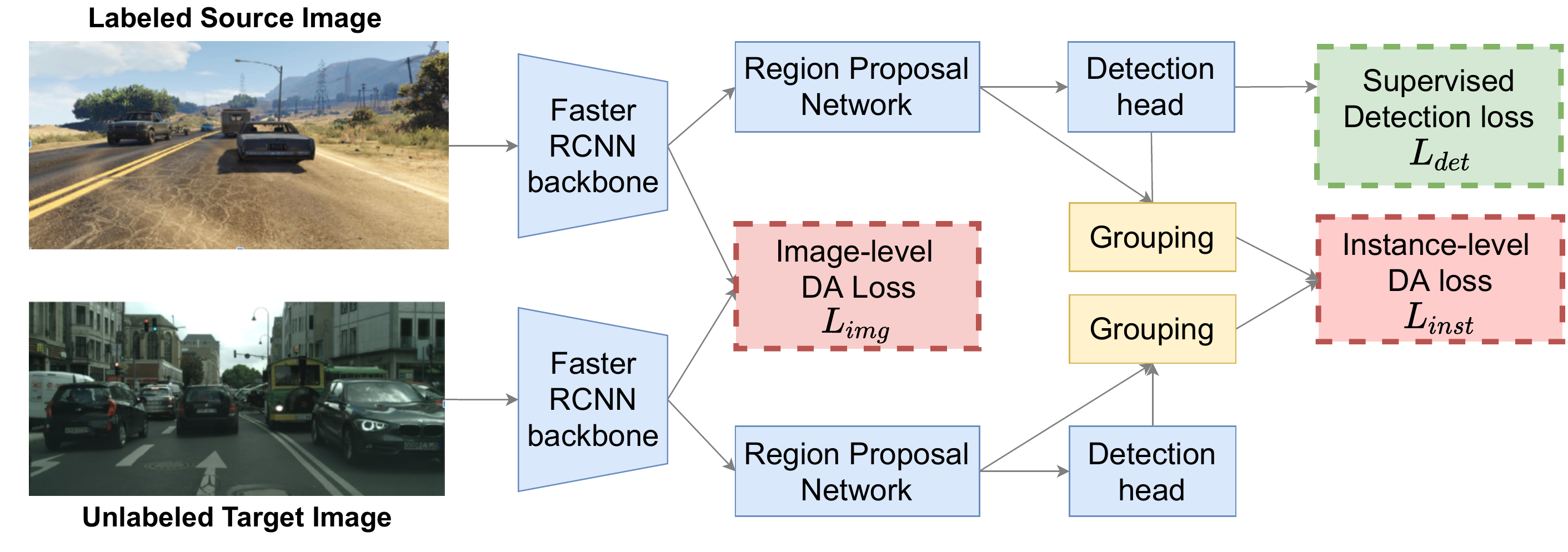}
	\caption{Components of our general unsupervised domain adaptation framework, for object detection. Here the boxes in blue are components of Faster R-CNN. They share parameters in both domains.
	} 
	\label{fig:overall}
	\vspace{-3mm}
\end{figure*}

In this section we discuss our general framework for analyzing several aspects of unsupervised domain adaptation methods for object detection.
%
Starting from the problem formulation, we present the main ingredients of our UDA framework (in~\ref{subsec:alignment_mechanism} and \ref{subsec:feature_levels}) which represent a generalization of the different components presented in the state-of-the-art. For each part, we discuss the existing alternatives that we later compare in \secref{subsec:res_experimental_analysis}.
%
%
We then introduce a novel algorithm (in~\ref{sec:visga}), \ourda, a direct implementation of our framework combining the best performing components 
with a novel strategy for a dynamic aggregation of  proposals based on their visual similarity.


%
\myparagraph{Problem formulation.} In Unsupervised Domain Adaptation (UDA) for object detection, we are given $N_\mathcal{S}$ labeled images for the source domain $\mathcal{S} = \{(x^\mathcal{S}_i, y^\mathcal{S}_i, B^\mathcal{S}_i)\}^{N_\mathcal{S}}_{i=1}$, where $y^\mathcal{S}_i$ and $B^\mathcal{S}_i$ are the class labels and bounding box coordinates respectively. For the target domain $\mathcal{T} = \{x^\mathcal{T}_i\}^{N_\mathcal{T}}_{i=1}$, only $N_\mathcal{T}$ unlabeled images are available.
Both domains share an identical label space but their visual distributions do not match. The goal of UDA approaches is to learn object detectors which perform well on the target domain, despite the domain shift.
\subsection{Overview} 
Our  generalized UDA framework comprises  of three main components.
First is a standard object detection network, Faster R-CNN, which takes an input image and produces bounding boxes and labels for all object instances present in the image. 
The second component is an image-level domain adaptation loss which encourages alignment of the global image representation in the backbone network.
The third component is an instance-level domain adaptation loss which induces alignment of representations of each object instance. 
This is illustrated in \figref{fig:overall}. 
Thus, the overall training objective of the method can be written as:
\begin{equation}
\mathcal{L} = \mathcal{L}_{det} + \lambda_1 \mathcal{L}_{img}+\lambda_2\mathcal{L}_{inst}, 
\label{eqn:final_loss}
\end{equation}
where, $\mathcal{L}_{det}$ is the supervised training loss for the detector, $\mathcal{L}_{img}$ and $\mathcal{L}_{inst}$ are the image-level and instance-level domain adaptation (DA) losses respectively, $\lambda_1$ and $\lambda_2$ are trade-off parameters. For methods that do not apply instant level alignment $\lambda_2$ is set to zero. Note that $\mathcal{L}_{det}$ is only applicable in the source domain where ground-truth bounding box annotations are available. 

\myparagraph{Detection network.} Following the convention set by early work on cross-domain object detection, we deploy Faster R-CNN \cite{da_faster_rcnn} as the object detection network in both, our method and the analysis. 
%
%
It consists of a Region Proposal Network (RPN) and a detection head. Both networks are trained with two loss terms each, a regression loss for bounding box estimation and a classification loss for label prediction. Thus the detection loss $\mathcal{L}_{det}$ for Faster R-CNN is composed of $\mathcal{L}_{RPN}$ and $\mathcal{L}_{head}$. 
\subsection{How to Induce Alignment?}
\label{subsec:alignment_mechanism}
The role of the domain adaptation losses ($\mathcal{L}_{img}, \mathcal{L}_{inst}$) is to induce alignment between the model's representation of source and target domain inputs. 
Downstream blocks that use such invariant representation (here for example RPN and the detection heads), 
 would be domain-agnostic and perform equally well in both domains.
While adversarial training has been the dominant paradigm for reducing the discrepancy between feature distributions~\cite{da_faster_rcnn, strong-weak, zhu_cvpr19_selective_alignment}, recently contrastive losses have been proposed to match source and target features \cite{GPA, kang2019contrastive}. We present these approaches in this subsection and compare them in our experimental analysis~(\secref{subsec:res_experimental_analysis}).

\myparagraph{Adversarial training.} 
The key idea in Adversarial Training~(AT) based UDA methods is to learn domain invariant representations by fooling a discriminator which is trained to predict the input data domain based on the detector features. 
This approach is usually class-agnostic, ignoring the features class information and focusing on domain-level alignment. 
Specifically, the features $F_d$ of domain $d$ ($d=0$ for source and $d=1$ for target) is fed to the discriminator $\mathcal{D}$ which predicts the domain of the extracted features. 
The discriminator is trained by minimizing the cross-entropy loss as below.
\begin{equation}
    \mathcal{L}_{disc} = -d\log(\mathcal{D}(F_{d})) - (1-d)\log(1- \mathcal{D}(F_{d})). 
    \label{eq:disc}
\end{equation}
Since we want to adapt the features of the two domains to be indistinguishable by the discriminator, we have to maximize the loss in \eqnsref{eq:disc} w.r.t the features $F_d$. This is achieved by incorporating a gradient reverse layer (GRL) \cite{grl_ganin}, before features are input to the discriminator.

\myparagraph{Contrastive learning.} 
As an alternative to AT, one can apply max-margin contrastive losses to align source and target features by leveraging the class information.
The main idea here is to push features from the same class closer and push apart features belonging to different classes across domains.
When matching a single feature vector $F_d^i$~per class $i$ in each domain $d \in (0,1)$, the max-margin contrastive loss takes the form: 
%
\small
\begin{equation} \label{eq:max_margin}
	\begin{split}
    \mathcal{L}_{CL} = \sum_i^C \left[ ||F_0^{i} - F_1^{i}||_2^2 + \sum_{j, j\neq i}^C max\{0, m - ||F_0^{i} - F_1^{j}||_2^2\}\right] 
	\end{split}
\end{equation}
\normalsize
where $C$ is the number of classes and $m$ is the margin.  
Since target data is unlabeled, the class prediction by the detector is used as a pseudo-label in \cite{GPA} to apply \eqnref{eq:max_margin}.
In our analysis, we also study the effect of ignoring this class information. 
This can be achieved by considering only two sets of vectors of cardinality $N_0$ and $N_1$, possibly unequal number ($N_0 \neq N_1$), from source and target domains to align.  
To apply contrastive losses here, we make a simple modification. 
Instead of matching class-specific features across domains, we match the proposals from one domain to the closest features (nn) of the other domain~(\ref{eq:max_margin_proposal_nn}) and minimize the distance between their representations~(\ref{eq:max_margin_proposal}), as shown below.
\begin{small}
	\begin{align} \label{eq:max_margin_proposal_nn}
        \text{nn}(i) &= \text{argmin}_{j<N_1} ||F_0^{i} - F_1^{j}|| \\
        \label{eq:max_margin_proposal}
        \begin{split}
            \mathcal{L}_{CL} &= \sum_i^{N_0} \big[ ||F_0^{i} - F_1^{\text{nn}(i)}||_2^2 
            \\&+ \sum_{j, j\neq \text{nn}(i)}^{N_1} max\{0, m - ||F_0^{i} - F_1^{j}||_2^2\} \big].
        \end{split}
	\end{align}
\end{small}
%
In our method, we utilize AT, avoiding  potential noise as a result of the reliance on unstable pseudo-labels during the alignment process. Our  aggregation strategy can leverage proposals similarities and possible embedded class-information as we explain in the sequel. 
\subsection{What Features to Align?}
\label{subsec:feature_levels}
In detection, two main levels of feature alignment can be considered: 1)  image-level features output by the backbone network and 2)  instance-level or object-level features obtained after pooling each region-of-interest proposed by the RPN network. 
%
%
%
The predominant approach aims for complete alignment at instance-level, i.e.\ the representation of every proposed object, in source or target domain, should be domain agnostic. 
This might be difficult to achieve, especially when complete alignment is challenging for the model, and when the source or target data during alignment contains some domain-specific outliers, e.g.\ specific backgrounds only found in a simulation domain. 
To address this, recent works aggregate the proposals on each of the source and target before applying feature alignment~\cite{GPA, zheng_cvpr20_prototype, zhu_cvpr19_selective_alignment}.
Both~\cite{GPA} and \cite{zheng_cvpr20_prototype} take it to the other extreme, by collapsing the instances into a single prototype per category.
While~\cite{GPA} merges prototypes based on spatial overlap using intersection-over-union (IoU) and class labels, \cite{zheng_cvpr20_prototype} only uses class labels to mean pool proposals into prototypes.
In contrast~\cite{zhu_cvpr19_selective_alignment} treads a middle ground by merging proposals into many discriminative regions, but still only using spatial overlap as the merging criteria.

In our analysis in \secref{subsec:res_experimental_analysis}, we compare the effectiveness of different components of this aggregation including 1) spatial grouping vs similarity based grouping (discussed in \secref{sec:visga}) 2) using class information vs class agnostic and 3) single prototypes vs multiple groups.
\subsection{Similarity-based Group Alignment} 
\label{sec:visga}
In this section, we propose a novel similarity-based grouping to aggregate object proposals before performing alignment.
%
%
We first aggregate proposals based on visual similarity into varying number of feature groups.
AT is then applied to align the mean embeddings of the groups extracted from the source and target domains. 
This simple yet effective change brings three key benefits. 
First, adversarial training at group level enables our model to coarsely align the main feature clusters, instead of attempting complete alignment of all instances which might be infeasible.
Second, in contrast to the spatial overlap used in \cite{GPA, zhu_cvpr19_selective_alignment}, visual similarity-based clustering allows our model to group objects which are located far away in the image, but look similar. 
Note that this still groups heavily overlapping proposals, since they tend to also be visually similar. Hence, it avoids producing duplicate visually similar groups.
%
By using visual similarity, we do not depend on the pseudo-labels different from  previous approaches~\cite{GPA, zheng_cvpr20_prototype}.  
The pseudo-labels tend to be noisy,   thus avoiding such dependency can be beneficial especially in early training.
Moreover, when similar proposals are aggregated, we can implicitly leverage class information since the aggregated proposals are likely to be of the same class. 
Finally, by adaptively varying the number of groups, instead of using single prototypes, our model retains sufficient capacity to represent intra-domain variance. 


%
\myparagraph{Similarity-based clustering.} To perform similarity-based clustering, we take as input the $N$ proposals generated by RPN and their fixed feature vectors denoted by $f\in \mathbb{R}^{N \times m}$.
In order to discover the main feature groups, we cluster these features using  hierarchical agglomerative clustering. Starting bottom-up, each proposal is considered as an individual cluster. Then, at each step, the two closest clusters according to a distance metric are merged together. We utilize cosine distance as our merging metric:
\begin{equation} \label{eq:metric}
distance (z_{i} , z_{j}) =1 - \frac{z_{i} . z_{j}}{||z_{i}|| \ ||z_{j}||}, 
\end{equation}
\noindent where $z_{i}$ and $z_{j}$ show $i$-th and $j$-th proposal's feature embeddings. In contrast to recent work~\cite{GPA}, which uses spatial overlap~(measure by IoU) to group together instances, using cosine similarity enables us to pair instances which are located far from each other, but are visually similar. 
Merging is stopped when dissimilarity within a cluster, as defined by a \emph{linkage function}, exceeds the cluster radius parameter $\tau$. We apply the complete-linkage heuristic~\cite{defays1977efficient}, which ensures that the farthest distance of two members is smaller than $\tau$.
\begin{equation} \label{eq:linkage}
\text{MaxLink}(A,B) = \ max \{ dist(a,b) : a\in A, b\in B\},
\end{equation}
\noindent where A, B are two sets of proposals' features in two clusters and $dist$ is the cosine distance. 
This hierarchical clustering approach allows our model to adaptively change the number of feature groups during training, instead of having a fixed number of cluster like in k-means. 
Once the clustering has converged, instances assigned to each cluster are pooled to construct a representative embedding $Z_{c_i}$:
\begin{equation} 
Z_{c_i} = \frac{\sum_{i=0}^{N_{c_i}} \, z_i}{N_{c_i}},
\end{equation}
\noindent where $N_{c_i}$ is the number of instances assigned to the cluster $c_i$. 
The group representative $Z_{c_i}$ is fed to a group-level discriminator and adversarial training is applied to align groups from the two domains using \eqnref{eq:disc}.

Finally, our method~(\ourda) combines image and instance-level alignment of aggregated proposals via adversarial training as illustrated in Fig.\ref{fig:teaser}.

%% file: sections/analysis.tex
Based on \secref{sec:components}, 
we conduct ablation studies to analyze these design choices in~(\ref{subsec:res_experimental_analysis}) 1) AT vs CL for inducing feature alignment and 2) different feature levels for alignment. Then we compare our method, that combines the best performing components with a novel similarity-based grouping strategy, to SOTA results in~(\ref{sec:comppriorwork}). 
First, we present the datasets and the baselines used in the remainder of the paper.

\subsection{Experimental Setup}
\label{sec:expsetup}
We now present the datasets used for the experiments in the three domain shift scenarios. 

\myparagraph{Adverse weather.} For this scenario, we use Cityscapes \cite{city} as the source dataset. It contains 3,475 real urban images, with 2,975 images used for training and 500 for the validation. Foggy version of Cityscapes \cite{foggy} is used as the target dataset. 
%
Highest fog intensity (least visibility) images are used in our experiments, matching prior work~\cite{GPA}. 
Both datasets have 8 different categories. 
Following \cite{da_faster_rcnn}, we used the tightest bounding box of an instance segmentation mask as ground truth box. This scenario is referred to as \emph{Foggy}.

\myparagraph{Synthetic to real.} SIM10k~\cite{sim10k} is a simulated dataset that contains  10,000 synthetic images. 
In this dataset, we use all 58,701 car bounding boxes available as the source data during training. For the target data and evaluation, we use Cityscapes~\cite{city} and only consider the car instances. This scenario is referred to as \emph{Sim2Real}.

\myparagraph{Cross camera.} In this scenario, we use real the dataset of KITTI~\cite{kitti} as our source data. KITTI contains 7,481 images and we use all of them for training. Similar to the previous scenarios, we use Cityscapes~\cite{city} as target data. 

In all experiments, we use mean average precision (mAP) with IoU threshold of 0.5 for evaluation. We compare our approach with the following prior works: DA \cite{da_faster_rcnn}, DivMatch \cite{diversify_and_match}, SW-DA \cite{strong-weak}, SC-DA \cite{zhu_cvpr19_selective_alignment} and MTOR \cite{mean_teacher}. 

%
%
\myparagraph{Implementation details.} We set the shorter side of the image to 600 pixels, following the Faster R-CNN implementation \cite{ren2015faster}. Our Faster R-CNN network, as well as all the prior works we compare to, utilize ResNet-50~\cite{he2016deep} as the backbone. Models using adversarial training are first trained with learning rate 0.001 for 50K iterations, then with learning rate 0.0001 for 20K more iterations and we report the final performance. Each batch is composed of 2 images, one from each domain. A momentum of 0.9 and a weight decay of 0.0005 is used. With the mentioned setting, maximum ~10k MB of memory needed and one NVIDIA Tesla V100-PCIE GPU is used. 
For training contrastive learning models, we employ the code provided by \cite{GPA} and we follow its exact settings for running experiments. Both methods are implemented with PyTorch~\cite{pytorch}.

\subsection{Analysis of UDA Components}
\label{subsec:res_experimental_analysis}
In this section we analyze the various design choices of alignment mechanisms~(\Tableref{ablation_cl_at}), image-level alignment~(\Tableref{tab:imglevel}), aggregation levels and aggregation mechanisms~(\Tableref{tab:ablation_foggy_sim}) when  bulding UDA models.
%

In Table~\ref{ablation_cl_at}, we compare CL and AT domain alignment paradigms in the \emph{Sim2Real} scenario.
Faster R-CNN is the baseline model which is only trained on the source and tested on the target. Single and Multiple Group(s) are shown as SG and MG. CA represents Class Agnostic, which means that class information is not used when constructing the groups. 
%
%
CL using SG as aggregation level, improves the performance over the source-only model ($33.2\%$ vs $31.9\%$).
Similarly, applying CL with MG ($36.9\%$) or MG+CA ($42.6\%$) setup further improves  model performance.
AT  outperforms CL in each of these three scenarios (fifth to seventh rows of \Tableref{ablation_cl_at}).
%
Applying AT on the SG results in a large improvements over the baseline ($40.8\%$ vs $31.9\%$).
%
Similarly, AT heavily outperforms CL for the MG setting ($43.1\%$ vs $36.9\%$). 
Same trend is observed in MG+CA as well, with AT outperforming CL ($45.6\%$ vs $42.6\%$). 
This large margin reveals that allowing the network to freely align the group representatives with AT, leads to a larger performance gain compared to explicitly matching the groups to nearest neighbors across domains using CL. 
%
Based on these results, we use AT for the rest of experiments.

\begin{table}
    \centering
    \scalebox{0.9}{
    \setlength{\tabcolsep}{5.0mm}
    \begin{tabular}{lcc}
    	\toprule[1.0pt]
    	 Method & Agg. Levels & \emph{car} AP \\
    	\cmidrule(lr){1-1}\cmidrule(lr){2-2}\cmidrule(lr){3-3}
    	Faster R-CNN & --- & 31.9\\\cmidrule(lr){1-1}\cmidrule(lr){2-2}\cmidrule(lr){3-3}
    	\multirow{3}{*}{\parbox{2.0cm}{Contrastive losses}} & SG & 33.2\\
    	& MG & 36.9 \\
    	& MG+CA & 42.6 \\
        \cmidrule(lr){1-1}\cmidrule(lr){2-2}\cmidrule(lr){3-3}
        \multirow{3}{*}{\parbox{2.0cm}{Adversarial training}} & SG & 40.8 \\
    	 & MG & 43.1 \\
    	 & MG+CA & 45.6 \\
    	\bottomrule[1.0pt]
    \end{tabular}}
    \vspace{1mm}
    \caption{\textbf{Sim2Real:} Analyzing the choice of alignment mechanism, comparing adversarial training against contrastive learning across different aggregation conditions (SG: Single Group, MG: Multiple Groups, CA: Class Agnostic ). Note that all results here only use instance level alignment.} 
    \label{ablation_cl_at}
\end{table}


\begin{figure*}
	\centering
	\setlength{\tabcolsep}{0.02mm}
	\begin{tabular}{cccc}
	 \includegraphics[valign = c,width=0.22\linewidth]{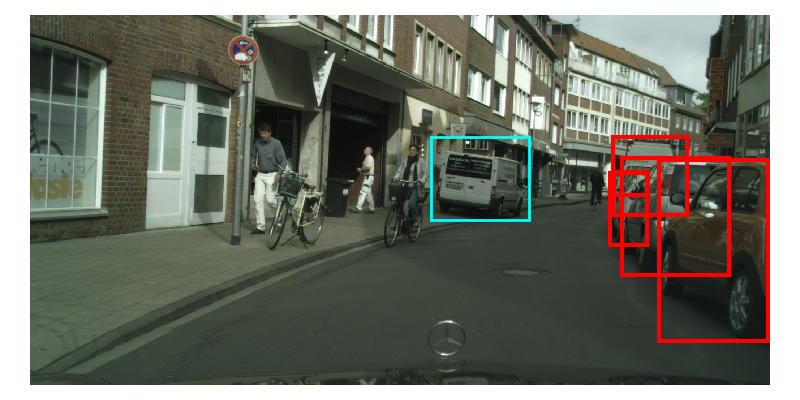}&
	 \includegraphics[valign = c,width=0.22\linewidth]{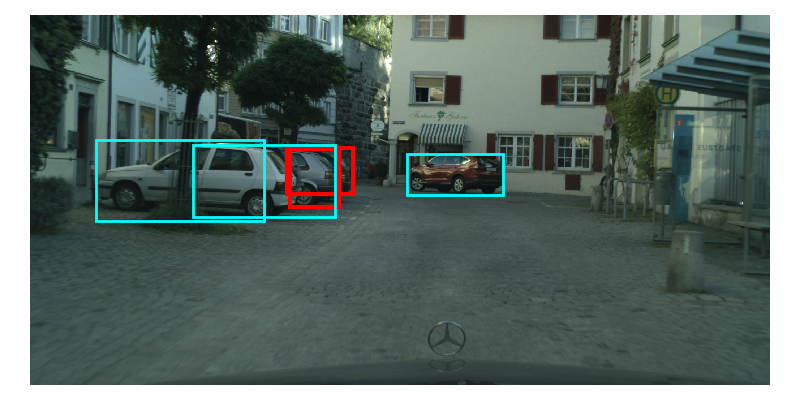}&
	 \includegraphics[valign = c,width=0.22\linewidth]{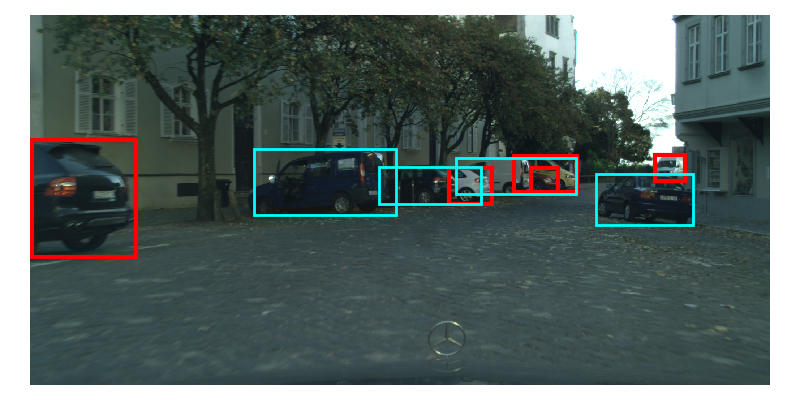}&
	 \includegraphics[valign = c,width=0.22\linewidth]{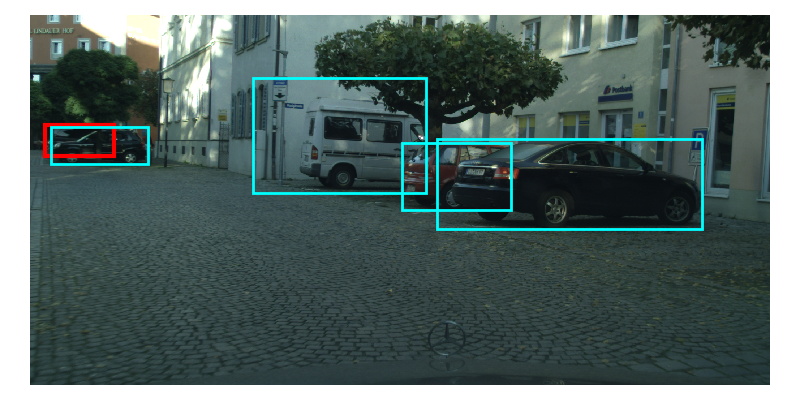}\\
	 \includegraphics[valign = c,width=0.22\linewidth]{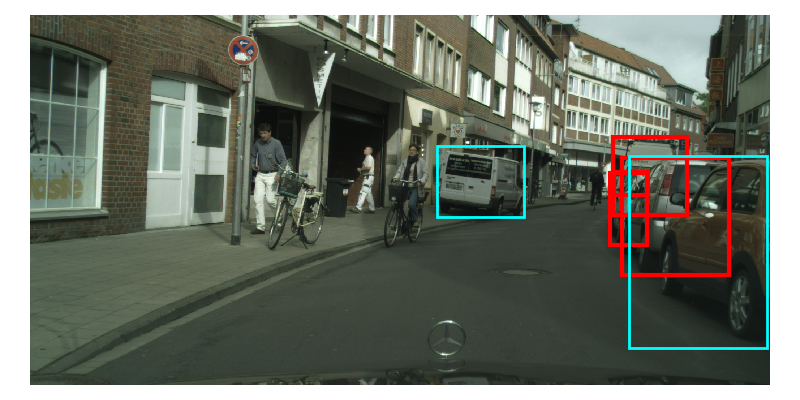}&
	 \includegraphics[valign = c,width=0.22\linewidth]{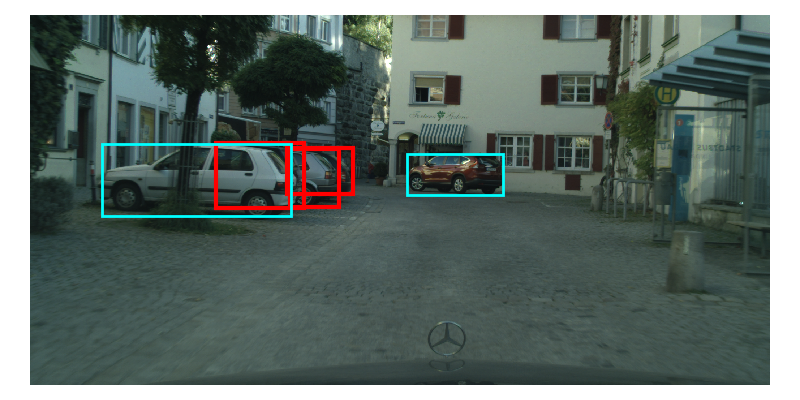}&
	 \includegraphics[valign = c,width=0.22\linewidth]{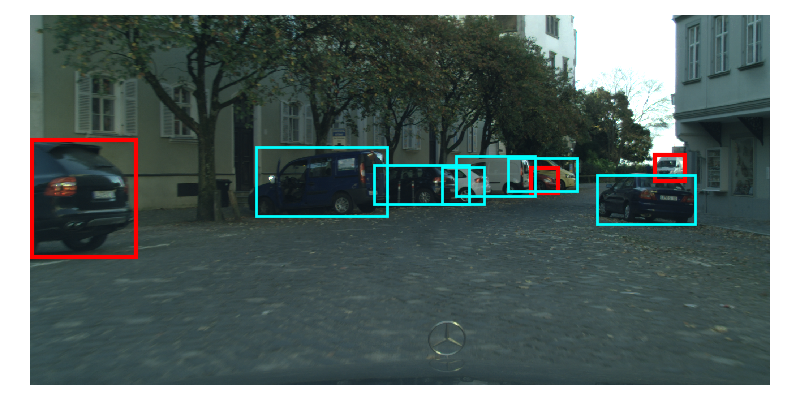}&
	 \includegraphics[valign = c,width=0.22\linewidth]{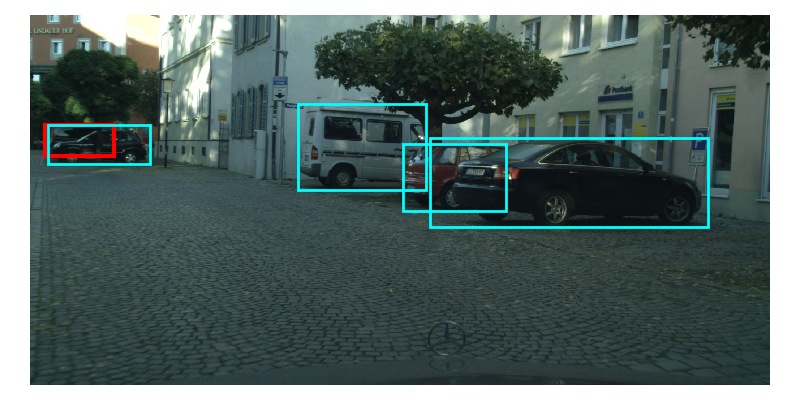}\\
	 \includegraphics[valign = c,width=0.22\linewidth]{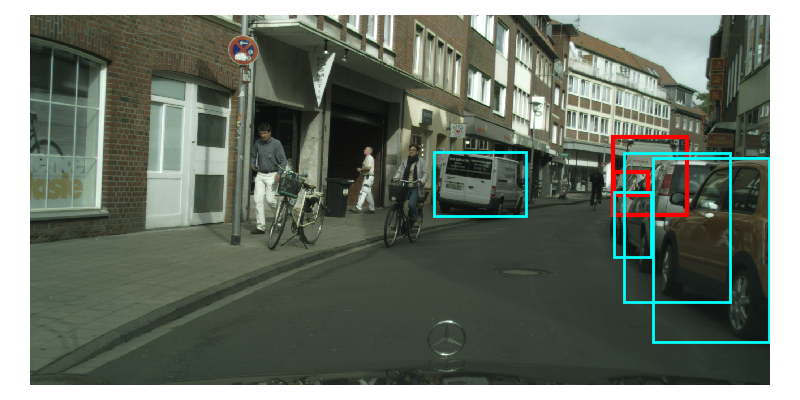}&
	 \includegraphics[valign = c,width=0.22\linewidth]{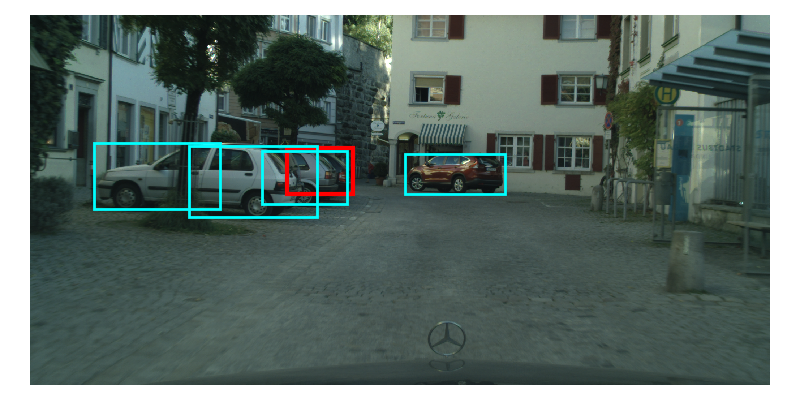}&
	 \includegraphics[valign = c,width=0.22\linewidth]{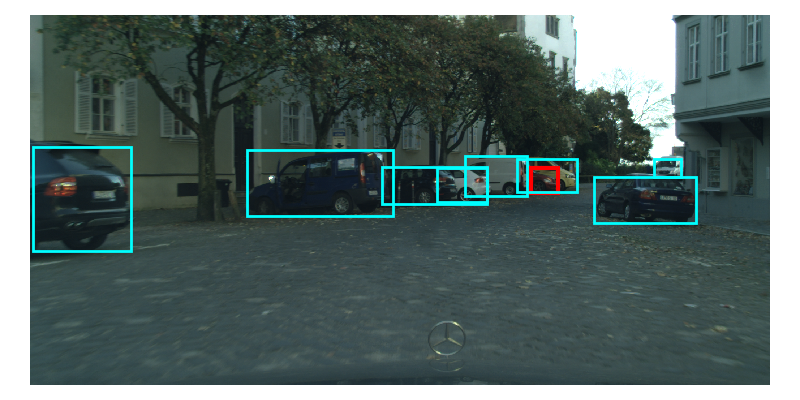}&
	 \includegraphics[valign = c,width=0.22\linewidth]{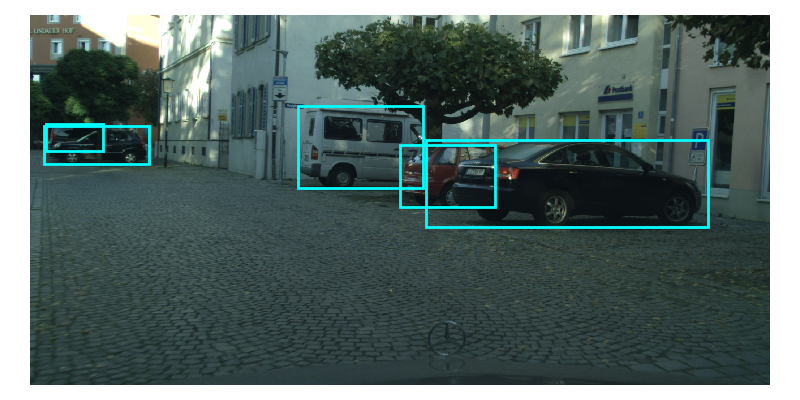}\\
	\end{tabular}
	\caption{Qualitative results. Sim2Real scenario. First row: Faster R-CNN, second row: ViSGA (iou) and, last row: ViSGA (cosine). True positives and missed objects are shown as cyan and red boxes respectively. We can clearly see that Faster R-CNN model misses many objects. This improves in the second row, with the model based on grouping proposals with spatial overlap. However, \ourda model powered by similarity-based aggregation does even better, recovering almost all missed objects.
	} 
	\vspace{0mm}
	\label{fig:qualitative_results}
\end{figure*}
\begin{table}
    \centering
    \scalebox{0.9}{
    \setlength{\tabcolsep}{4.0mm}
    \begin{tabular}{cccc}
    	\toprule[1.0pt]
    	 Image-level & SG & MG & MG+CA\\
    	\cmidrule(lr){1-1}\cmidrule(lr){2-4}
    	\xmark & \bf40.8& 43.1& 45.6\\
    	\cmark & 39.5& \bf 44.9& \bf 49.3\\
    	\bottomrule[1.0pt]
    \end{tabular}}
    \vspace{1mm}
    \caption{\textbf{Sim2Real:} Analyzing the effect of image-level alignment.} 
    \label{tab:imglevel}
    \vspace{0mm}
\end{table}
\myparagraph{Do we need image-level alignment?} \Tableref{tab:imglevel} presents the comparison of model performance with image-level alignment added on top of the instance-level alignment presented before. 
This comparison is done using AT across different aggregation levels. 
We see that image-level alignment brings clear added improvement on both multi-group models, while degrading slightly on the single group model. 
On the single group model, the instance-level alignment is happening at a global level since all the instances are aggregated into a single group before inducing alignment. Adding an extra alignment will not help further and could possibly induce noise, as seen in the results ($40.8\%$ vs $39.5\%$). However, on models with multiple groups, instance level AT focuses on local feature alignment, and hence adding global alignment with image-level AT is beneficial. Thus, we use image-level alignment for the remaining of the experimental section.

\begin{table}
     \centering
     \scalebox{0.9}{
	\begin{tabular}{*{2}{c} cc}
		\toprule[1.0pt]
		Aggr.\ levels & Aggr.\ mechanism &  Foggy& Sim2Real \\\cmidrule(lr){1-1}\cmidrule(lr){2-2}\cmidrule(lr){3-4} 
		Proposals & No grouping & 38.5  & 39.0 \\\cmidrule(lr){1-1}\cmidrule(lr){2-2}\cmidrule(lr){3-4}
		SG & \multirow{4}{*}{Cosine} & 33.7 & 39.5 \\
		MG (adaptive) &  & 41.8  & 44.9 \\
		MG+CA (adaptive) & & \textbf{43.3} & \textbf{49.3}\\
		MG+CA (fixed) & & 42.5 & 49.0
		 \\\cmidrule(lr){1-1}\cmidrule(lr){2-2}\cmidrule(lr){3-4}
		MG+CA (adaptive) & IoU & 41.9 & 44.8\\
		\bottomrule[1.0pt]
	\end{tabular}}
	\vspace{1mm}
	\caption{
	\textbf{Sim2Real \& Foggy:} Analyzing the choice of different aggregation levels and mechanisms.}
    \label{tab:ablation_foggy_sim}
    \vspace{0mm}
\end{table}

\myparagraph{Aggregation levels \& mechanisms.} Next, we study the process of aggregating instance proposals into groups before performing alignment. We compare the effect of both the number of groups as well as the mechanism used to aggregate proposals into groups. Table~\ref{tab:ablation_foggy_sim}, 
%
first row shows results using original proposals without any grouping for the instance level alignment. 
Aggregating instances into a SG per category causes a significant drop in performance, indicating that the condensing features into one vector may not be a useful approach. 
However, MG setup based on visual similarity (Cosine), is beneficial ($41.8\%$ vs $38.5\%$ on \emph{Foggy} and $44.9\%$ vs $39.0\%$ on \emph{Sim2real} ).
Performance is further improved by ignoring the predicted class-label ($43.3\%$ on \emph{Foggy} and $49.3\%$ on \emph{Sim2real}) and compared to the last set with MG, this shows that noisy pseudo labels (in MG) can be harmful to the clustering process and may have negative impact on the alignment.
%
%
Both the above models use cluster radius parameter to let the model vary the number of groups adaptively over the course of training.
Here, we do not compare different clustering methods directly. However, we also experiment with fixing the number of clusters, as shown in \emph{MG+CA (fixed)}. we perform a sweep of the number of clusters hyper-parameter and report the best numbers here (full results can be found in supplementary, figure 4). 
This model performs slightly worse than \emph{MG+CA (adaptive)}, indicating that the flexibility from adaptive number of clusters is beneficial.

Finally in the last row, 
by using spatial overlap~(using IoU) to cluster instances~(as proposed in~\cite{GPA, zheng_cvpr20_prototype}), we see that the performance drops by 1.4\% and 4.5\% on \emph{Foggy} and \emph{Sim2Real} respectively, compared to using visual similarity based clustering~(MG+CA (adaptive)). These large drops show that our visual similarity based grouping is a better way to accumulates proposals, since it allows grouping distant instances and avoids redundant group representatives. 

\subsection{Comparison with SOTA}
\label{sec:comppriorwork}

In this section, we evaluate the best design choices embedded in our \ourda  
and compare it to prior works in each of these domains in \secref{sec:related}.
\ourda incorporates image-level alignment and adversarial training framework along with the novel group alignment of visual similarity based class-agnostic clusters.
\begin{table}[t]
     \centering
     \scalebox{0.8}{
	\begin{tabular}{{c}cc}
		\toprule[1.0pt]
		Methods &  Cross Camera & Sim2Real \\\cmidrule(lr){1-1}\cmidrule(lr){2-3} 
		Faster R-CNN  & 32.5  & 31.9 \\\cmidrule(lr){1-1}\cmidrule(lr){2-3}
		DA-Faster \cite{da_faster_rcnn} & 41.8 & 41.9 \\
		DivMatch \cite{diversify_and_match} & 42.7  & 43.9 \\
		SW-DA \cite{strong-weak} & 43.2 & 44.6\\
		SC-DA \cite{zhu_cvpr19_selective_alignment} & 43.6 & 45.1\\
		MTOR \cite{mean_teacher} & - & 46.6\\
		GPA (Only RCNN) \cite{GPA} & 46.1 & 44.8\\
		GPA \cite{GPA} & 47.9 & 47.6\\
		\cmidrule(lr){1-1}\cmidrule(lr){2-3}
		Ours & 47.6 & \textbf{49.3}\\
		\bottomrule[1.0pt]
	\end{tabular}
	}
	\vspace{1mm}
	\caption{Experimental results (\%) of \emph{Sim2Real \& Cross Camera}.} \label{eval_sim10k_kitti_city}
\end{table}



%
Table~\ref{eval_sim10k_kitti_city} shows the results for the \emph{Sim2Real} and \emph{Cross Camera} scenarios on \emph{Car} class. 
The adaption is challenging on \emph{Sim2Real} due to relatively large domain shift between source and target. However, as shown in the table, our approach outperforms other methods by a fair margin ($49.3\%$ vs $47.6\%$ by the closest model, GPA). 
For Cross Camera scenario, our approach has competitive performance compared to GPA\cite{GPA}, while out-performing other approaches. 
In Table~\ref{eval_foggy_cs}, \ourda achieves SOTA results, with large improvements over other recent work. 
It outperforms the GPA method~\cite{GPA} ($43.3\%$ vs $39.5\%$) based on prototype matching, highlighting the importance of our design choices --- multiple similarity based class-agnostic groups and adversarial training.
In summary, the good performance shown by our model across three datasets with state-of-the-art results in two of them, provides evidence that our similarity-based method is successful in aligning instance level representations. 
\begin{table}[]
\center
\scalebox{0.85}{
\tabcolsep=1.5pt
\begin{tabular}{ c | c c c c c c c c | c}
  \hlineB{3}
   Methods & prsn & rider & car & truck & bus & train & mcycle & bicycle & mAP \\ \hline
   Faster R-CNN & 27.2 & 31.8 & 32.5 & 16.0 & 25.5 & 5.6 & 19.9 & 27 & 22.8\\\hline
  DA-Faster \cite{da_faster_rcnn} & 29.2 & 40.4 & 43.4 & 19.7 & 38.3 & 28.5 & 23.7 & 32.7 & 32.0 \\
  DivMatch \cite{diversify_and_match} & 31.8 & 40.5 & 51.0 & 20.9 & 41.8 & 34.3 & 26.6 & 32.4 & 34.9 \\
  SW-DA \cite{strong-weak} & 31.8 & 44.3 & 48.9 & 21.0 & 43.8 & 28.0 & 28.9 & 35.8 & 35.3 \\
  SC-DA \cite{zhu_cvpr19_selective_alignment} & 33.8 & 42.1 & 52.1 & 26.8 & 42.5 & 26.5 & 29.2 & 34.5 & 35.9 \\
  MTOR \cite{mean_teacher} & 30.6 & 41.4 & 44.0 & 21.9 & 38.6 & 40.6 & 28.3 & 35.6 & 35.1 \\
  GPA \cite{GPA}  & 32.9 & \textbf{46.7} & 54.1 & 24.7 & 45.7 & 41.1 & \textbf{32.4} & 38.7 & 39.5 \\ \hline
    Ours & \textbf{38.8} &  45.9 & \textbf{57.2}  & \textbf{29.9} & \textbf{50.2} & \textbf{51.9} &  31.9 & \textbf{40.9}  & \textbf{43.3}\\
  \bottomrule[1.0pt]
\end{tabular}}
\vspace{1mm}
\caption{Experimental results of (\%) \emph{Foggy}.}
\vspace{0mm}
\label{eval_foggy_cs}
\end{table}

\begin{figure}[t]
	\centering
	\includegraphics[width=0.30\textwidth]{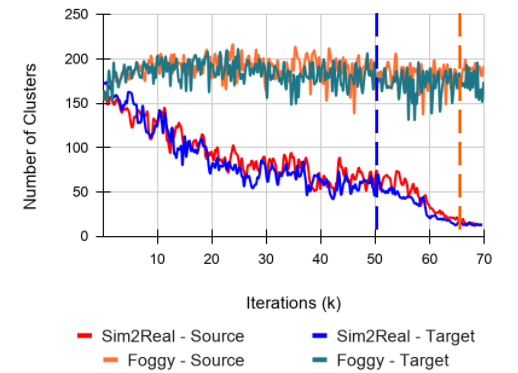}
	\caption{Evolution of number of groups during training for our \ourda model. Orange and blue dashed lines show the best training stops for Foggy and Sim2Real respectively. } 
	\label{fig:cluster_analysis}
	\vspace{0mm}
\end{figure}

\begin{figure}
	\centering
	\includegraphics[width=0.35\textwidth]{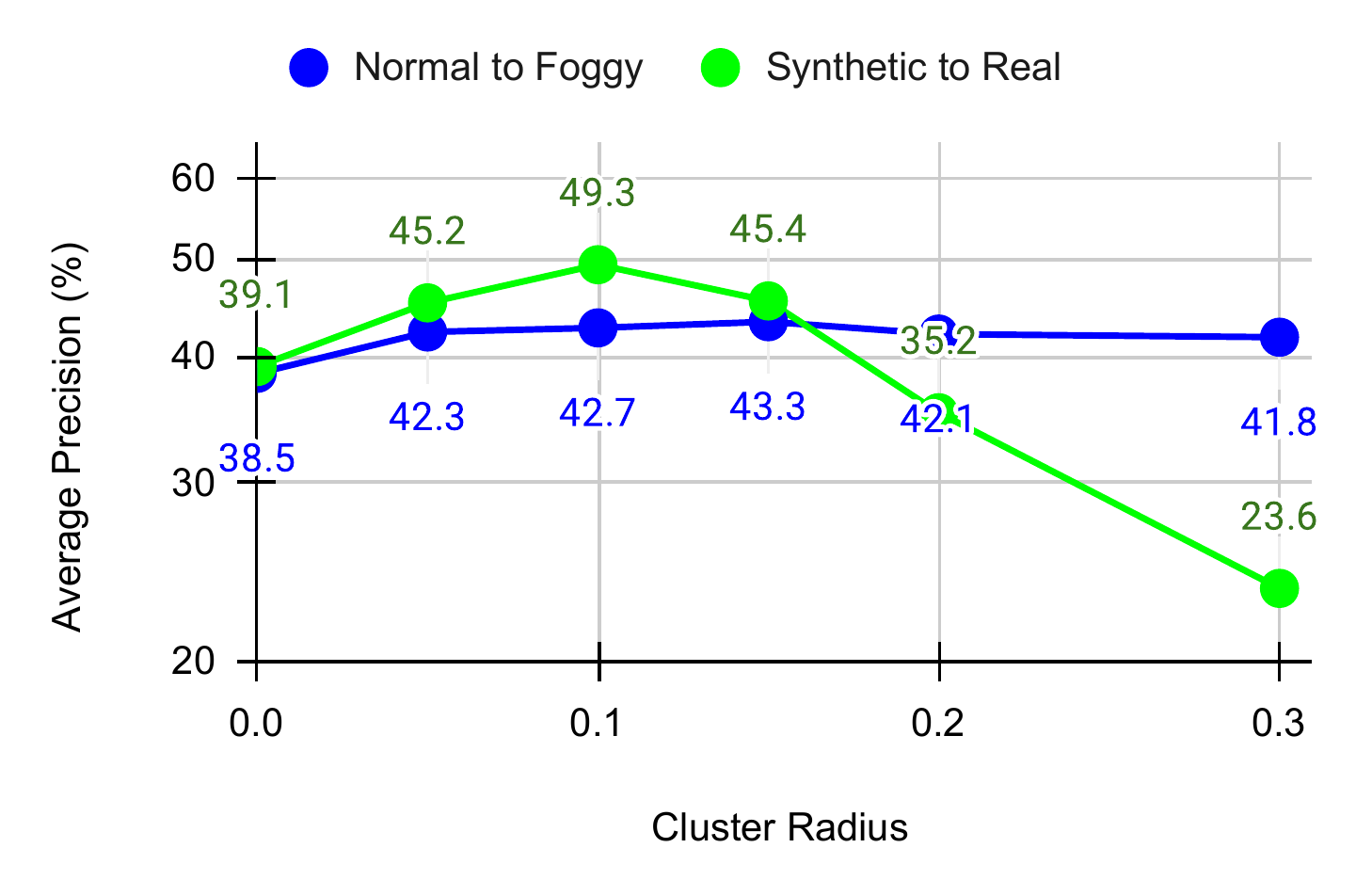}
	\caption{Sensitivity analysis of cluster radius parameter.}
	\label{sensitivity_analysis}
	\vspace{-2mm}
\end{figure}

\myparagraph{Qualitative analysis of \ourda.} 
%
\Figref{fig:qualitative_results} compares the detection outputs of Faster R-CNN and \ourda models with different aggregation mechanisms, on \emph{Sim2Real} scenario. 
%
%
\Figref{fig:cluster_analysis} shows the evolution of the number of groups during \ourda training, on \emph{Foggy} and \emph{Sim2Real}.
While the number of initial groups are similar in both cases, the number of clusters on Sim2Real drops-off quickly and settles around 50 clusters when the best model performance is achieved.
In contrast, in \emph{Foggy}, the number of clusters increases and is plateaus around 180, where the best performance is achieved. 
This difference can be understood by noting that the \emph{Foggy} scenario has 8 categories compared to only one category in Sim2real. 
Hence the model needs more clusters in \emph{Foggy}.
In figure~\ref{sensitivity_analysis} shows experimental results measuring the sensitivity of the cluster radius parameter. 
We can observe that for \emph{Sim2Real} the network performs well when the threshold is low but it is relatively sensitive to high or very low radius values (no grouping). 
This might be due to the large shift between synthetic images and real images. 
In addition, a low cluster radius creates many single member clusters, reducing information aggregation. 
In contrast, the performance is not very sensitive to various radius values on \emph{Foggy}, where the domain gap is smaller. 
%
%
Additionally, figure 6 in the supplement presents a tSNE  \cite{van2008visualizing} visualizations of source and target feature distribution, to visually illustrate how \ourda prioritizes foreground alignment. This is also supported by figure 4 in the supplement, which shows that foreground objects get allocated more clusters and hence are prioritized for alignment.

\myparagraph{Computational Overhead.} The extra training time cost of our method, from computing the distances between the features of each proposal, is relatively small (eg.  one batch runtime is 0.79 for \ourda compared 0.62 for w/o \ourda). \ourda has no overhead during inference. Note that contrastive learning based methods, e.g., GPA, also compute the distance between proposals in each domain.

%% file: sections/generalization.tex

As mentioned in section~(\ref{sec:related}), existing UDA detector methods focus only on the single-source in which training data are gathered from one input domain guaranteeing homogeneity withing the training data. However, in real world annotated data is available or could be gathered under different conditions comprising different input domains, a scenario usually referred to as multi-source domain adaptation. %
In this section, we examine the applicability of our framework to operate in a multi-source UDA scenario. For the presented experiments, SIM10k and KITTI are used as source datasets and Cityscapes as the target. 

In the first set of experiment, we combine all sources into one training dataset and use shared discriminators at both image and instance levels for all different sources~(Figure~\ref{fig:ablation_source_combined}, part~(a)). 
 We repeat the same analysis, carried on the single source setting, to examine the right level of aggregation. 
As shown in Table~\ref{tab:eval_source_combined}, learning on multi-source data without any UDA components achieves $42.5\%$.  When combining the image and instance-level alignments it reaches to $49.6\%$~(Proposals). 
Using our full \ourda method, we can further improve the performance on multi-source~($51.3\%$). This confirms our method's scalability to multi-source setting. 
%
%
We also perform an ablation regarding the discriminator deployed in AT and modify the network design by 
considering a separate set of discriminators for each pair of source-target~(Fig.~\ref{fig:ablation_source_combined}, b to d, illustrates the different combinations).  
%
As we can see in~(Fig.~\ref{fig:ablation_source_combined}, e), our simple yet effective method with shared discriminators brings the largest gain to the final detection performance ($51.3\%$) compared to $50.0\%$ with separate discriminators at both instance and image level.

In summary, the good performance shown here provides further evidence that our method is able to generalize to multi-source setting without applying any modifications in its design.  This  leaves the door open for  exploring any alternatives that could further leverage the multi-source information in UDA object detection. 
%
\begin{table}
    \centering
    \scalebox{0.9}{
    \setlength{\tabcolsep}{1mm}
    \begin{tabular}{cccccc}
    	\toprule[1.0pt]
    	 Source & Faster & Proposals & SG & MG & MG+CA\\
    	\cmidrule(lr){1-1}\cmidrule(lr){2-3}\cmidrule(lr){4-6}
    	Single-Source (KITTI) & 32.5 & 41.5 & 35.8 & 45.5 & 47.6 \\
    	Single-Source (SIM10k) & 31.9 & 39.5 & 39.5 & 44.9 & 49.3\\
    	Multi-Source & 42.5 & 49.6 & 48.9 & \bf 51.3 & \bf 51.3\\
    	\bottomrule[1.0pt]
    \end{tabular}}
    \vspace{1mm}
    \caption{Multi-Source \ourda vs Single-Source \ourda. `Faster': No UDA; `Proposals': UDA with proposal-level alignment; `SG',`MG',`MG+CA': UDA with group-level alignment.}
    \label{tab:eval_source_combined}
    \vspace{0mm}
\end{table}

\begin{figure}
	\centering
	\setlength{\tabcolsep}{.5mm}
	\begin{tabular}{cc}
	 \includegraphics[valign = c,width=0.45\linewidth]{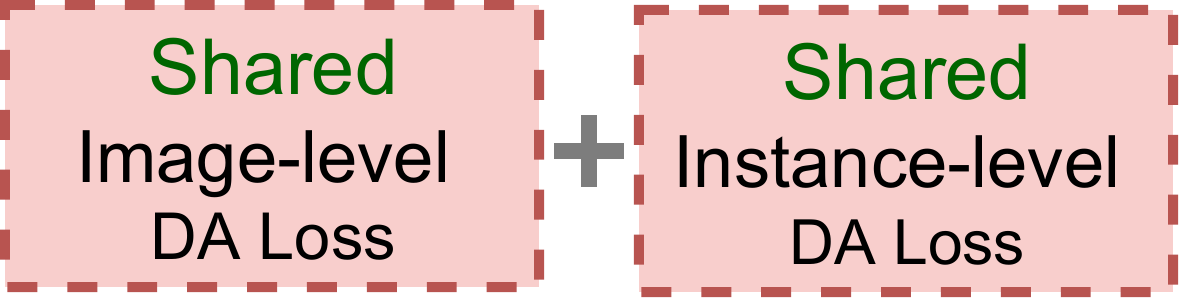}&
	 \includegraphics[valign = c,width=0.45\linewidth]{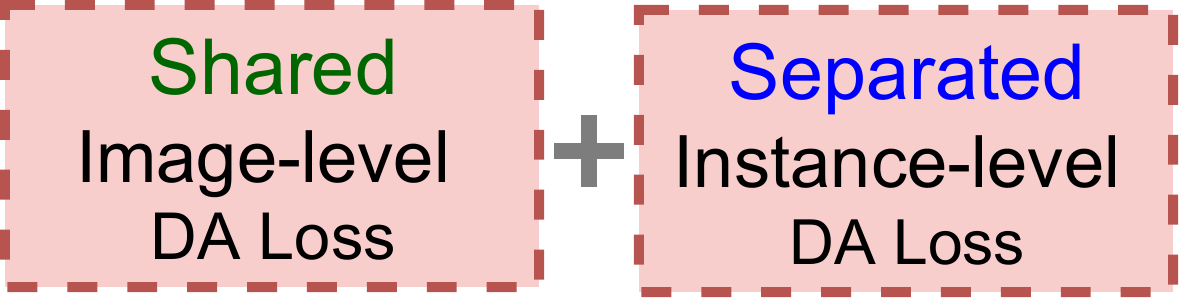}\vspace{1mm}\\
	 (a)  & (b) \vspace{1mm}\\
	 \includegraphics[valign = c,width=0.45\linewidth]{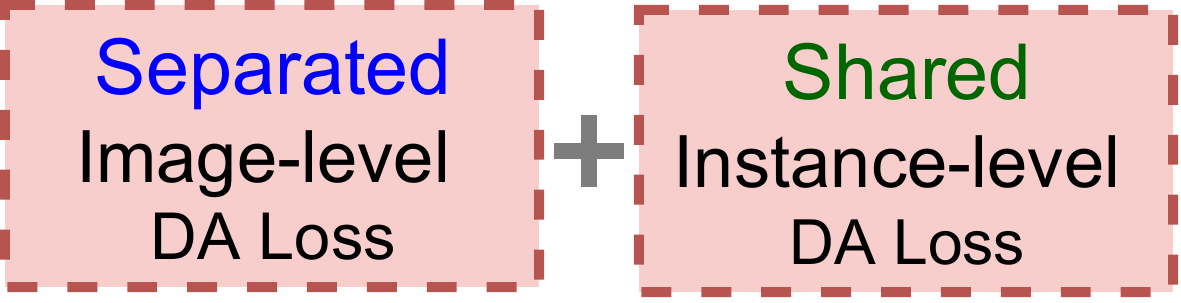}&
	 \includegraphics[valign = c,width=0.45\linewidth]{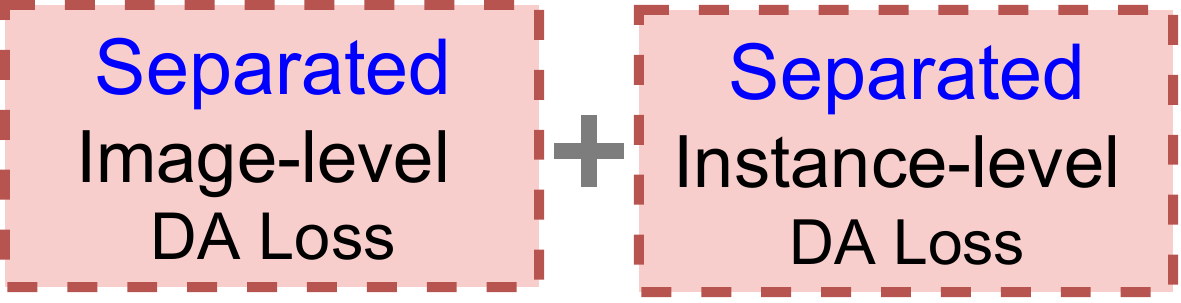}\vspace{1mm}\\
	 (c) & (d)\vspace{1mm}\\
	\end{tabular}
	\scalebox{0.85}{
    \begin{tabular}{ccccc}
    	\toprule[1.0pt]
    	 Model & Shared & Img & Ins & Separated\\
    	\cmidrule(lr){1-1}\cmidrule(lr){2-5}
        Multi-Source & \bf 51.3 & 49.1 & 49.3 & 50.0 \\
    	\bottomrule[1.0pt]
    \end{tabular}}
    \\(e)\\
	\caption{Multi-Source \ourda Ablation:  Shared/Separated discriminators between sources. (a) Shared.\ (b) Ins: separated image-level disc.\ (c) Img: separated instance-level disc.\ (d) Separated. (e) Results on (a to d).}
	\label{fig:ablation_source_combined}
	\vspace{0mm}
\end{figure}

%% file: sections/conclusions.tex
We present an analysis of various design choices when building UDA models for detection. Our experiments comparing the alignment mechanisms revealed that adversarial training works better than max-margin contrastive losses across different feature aggregation-levels. Regarding instance-level alignment, our analysis shows that aggregating proposals into multiple visually similar groups before alignment is beneficial. It significantly outperforms both options previously investigated in prior work; no aggregation \cite{da_faster_rcnn} or collapsing everything to a single category prototype vector \cite{GPA, zheng_cvpr20_prototype}. We also show that constructing these groups without considering pseudo labels improves performance in single-source setting. 
Our best model \ourda, incorporating adversarial training and visual class-agnostic group not only achieves SOTA results on \emph{Sim2Real} and \emph{Foggy}, it also generalizes to multi-source. 

%% file: egpaper_final.bbl
\begin{thebibliography}{10}\itemsep=-1pt

\bibitem{mean_teacher}
Qi Cai, Yingwei Pan, Chong{-}Wah Ngo, Xinmei Tian, Lingyu Duan, and Ting Yao.
\newblock Exploring object relation in mean teacher for cross-domain detection.
\newblock In {\em {IEEE} Conference on Computer Vision and Pattern
  Recognition}, 2019.

\bibitem{cariucci2017autodial}
Fabio~Maria Cariucci, Lorenzo Porzi, Barbara Caputo, Elisa Ricci, and
  Samuel~Rota Bulo.
\newblock Autodial: Automatic domain alignment layers.
\newblock In {\em 2017 IEEE International Conference on Computer Vision
  (ICCV)}, pages 5077--5085. IEEE, 2017.

\bibitem{chen_cvpr20_htcn}
Chaoqi Chen, Zebiao Zheng, Xinghao Ding, Yue Huang, and Qi Dou.
\newblock Harmonizing transferability and discriminability for adapting object
  detectors.
\newblock In {\em Proceedings of the IEEE/CVF Conference on Computer Vision and
  Pattern Recognition}, pages 8869--8878, 2020.

\bibitem{Chen2015}
Xinlei Chen, Tsung-Yi~Lin Hao~Fang, Ramakrishna Vedantam, Saurabh Gupta, Piotr
  Dollár, and C.~Lawrence Zitnick.
\newblock Microsoft {COCO} captions: Data collection and evaluation server.
\newblock {\em arXiv preprint arxiv:1504.00325}, 2015.

\bibitem{da_faster_rcnn}
Yuhua Chen, Wen Li, Christos Sakaridis, Dengxin Dai, and Luc~Van Gool.
\newblock Domain adaptive faster {R-CNN} for object detection in the wild.
\newblock In {\em {IEEE} Conference on Computer Vision and Pattern
  Recognition}, 2018.

\bibitem{city}
Marius Cordts, Mohamed Omran, Sebastian Ramos, Timo Rehfeld, Markus Enzweiler,
  Rodrigo Benenson, Uwe Franke, Stefan Roth, and Bernt Schiele.
\newblock The cityscapes dataset for semantic urban scene understanding.
\newblock In {\em {IEEE} Conference on Computer Vision and Pattern
  Recognition}, 2016.

\bibitem{dalal2005histograms}
Navneet Dalal and Bill Triggs.
\newblock Histograms of oriented gradients for human detection.
\newblock In {\em 2005 IEEE computer society conference on computer vision and
  pattern recognition (CVPR'05)}, volume~1, pages 886--893. IEEE, 2005.

\bibitem{defays1977efficient}
Daniel Defays.
\newblock An efficient algorithm for a complete link method.
\newblock {\em The Computer Journal}, 20(4):364--366, 1977.

\bibitem{duan2012domain}
Lixin Duan, Ivor~W Tsang, and Dong Xu.
\newblock Domain transfer multiple kernel learning.
\newblock {\em IEEE Transactions on Pattern Analysis and Machine Intelligence},
  34(3):465--479, 2012.

\bibitem{duan2011visual}
Lixin Duan, Dong Xu, Ivor Wai-Hung Tsang, and Jiebo Luo.
\newblock Visual event recognition in videos by learning from web data.
\newblock {\em IEEE Transactions on Pattern Analysis and Machine Intelligence},
  34(9):1667--1680, 2011.

\bibitem{pascal}
M. Everingham, L. Van~Gool, C.~K.~I. Williams, J. Winn, and A. Zisserman.
\newblock The pascal visual object classes (voc) challenge.
\newblock {\em International Journal of Computer Vision}, 88(2):303--338, June
  2010.

\bibitem{felzenszwalb2009object}
Pedro~F Felzenszwalb, Ross~B Girshick, David McAllester, and Deva Ramanan.
\newblock Object detection with discriminatively trained part-based models.
\newblock {\em IEEE transactions on pattern analysis and machine intelligence},
  32(9):1627--1645, 2009.

\bibitem{grl_ganin}
Yaroslav Ganin and Victor Lempitsky.
\newblock Unsupervised domain adaptation by backpropagation.
\newblock In {\em International conference on machine learning}, pages
  1180--1189. PMLR, 2015.

\bibitem{kitti}
Andreas Geiger, Philip Lenz, and Raquel Urtasun.
\newblock Are we ready for autonomous driving? the {KITTI} vision benchmark
  suite.
\newblock In {\em {IEEE} Conference on Computer Vision and Pattern
  Recognition}, 2012.

\bibitem{girshick2015fast}
Ross Girshick.
\newblock Fast r-cnn.
\newblock In {\em Proceedings of the IEEE international conference on computer
  vision}, pages 1440--1448, 2015.

\bibitem{girshick2014rich}
Ross Girshick, Jeff Donahue, Trevor Darrell, and Jitendra Malik.
\newblock Rich feature hierarchies for accurate object detection and semantic
  segmentation.
\newblock In {\em Proceedings of the IEEE conference on computer vision and
  pattern recognition}, pages 580--587, 2014.

\bibitem{gopalan2011domain}
Raghuraman Gopalan, Ruonan Li, and Rama Chellappa.
\newblock Domain adaptation for object recognition: An unsupervised approach.
\newblock In {\em 2011 international conference on computer vision}, pages
  999--1006. IEEE, 2011.

\bibitem{he2015spatial}
Kaiming He, Xiangyu Zhang, Shaoqing Ren, and Jian Sun.
\newblock Spatial pyramid pooling in deep convolutional networks for visual
  recognition.
\newblock {\em IEEE transactions on pattern analysis and machine intelligence},
  37(9):1904--1916, 2015.

\bibitem{he2016deep}
Kaiming He, Xiangyu Zhang, Shaoqing Ren, and Jian Sun.
\newblock Deep residual learning for image recognition.
\newblock In {\em Proceedings of the IEEE conference on computer vision and
  pattern recognition}, pages 770--778, 2016.

\bibitem{he_iccv19_MAF}
Zhenwei He and Lei Zhang.
\newblock Multi-adversarial faster-rcnn for unrestricted object detection.
\newblock In {\em Proceedings of the IEEE International Conference on Computer
  Vision}, pages 6668--6677, 2019.

\bibitem{sim10k}
Matthew Johnson{-}Roberson, Charles Barto, Rounak Mehta, Sharath~Nittur
  Sridhar, Karl Rosaen, and Ram Vasudevan.
\newblock Driving in the matrix: Can virtual worlds replace human-generated
  annotations for real world tasks?
\newblock In {\em {IEEE} International Conference on Robotics and Automation},
  2017.

\bibitem{kang2019contrastive}
Guoliang Kang, Lu Jiang, Yi Yang, and Alexander~G Hauptmann.
\newblock Contrastive adaptation network for unsupervised domain adaptation.
\newblock In {\em Proceedings of the IEEE Conference on Computer Vision and
  Pattern Recognition}, pages 4893--4902, 2019.

\bibitem{diversify_and_match}
Taekyung Kim, Minki Jeong, Seunghyeon Kim, Seokeon Choi, and Changick Kim.
\newblock Diversify and match: {A} domain adaptive representation learning
  paradigm for object detection.
\newblock In {\em {IEEE} Conference on Computer Vision and Pattern
  Recognition}, 2019.

\bibitem{krizhevsky2017imagenet}
Alex Krizhevsky, Ilya Sutskever, and Geoffrey~E Hinton.
\newblock Imagenet classification with deep convolutional neural networks.
\newblock {\em Communications of the ACM}, 60(6):84--90, 2017.

\bibitem{liu2016ssd}
Wei Liu, Dragomir Anguelov, Dumitru Erhan, Christian Szegedy, Scott Reed,
  Cheng-Yang Fu, and Alexander~C Berg.
\newblock Ssd: Single shot multibox detector.
\newblock In {\em European conference on computer vision}, pages 21--37.
  Springer, 2016.

\bibitem{long2015learning}
Mingsheng Long, Yue Cao, Jianmin Wang, and Michael Jordan.
\newblock Learning transferable features with deep adaptation networks.
\newblock In {\em International conference on machine learning}, pages 97--105.
  PMLR, 2015.

\bibitem{long2017deep}
Mingsheng Long, Han Zhu, Jianmin Wang, and Michael~I Jordan.
\newblock Deep transfer learning with joint adaptation networks.
\newblock In {\em International conference on machine learning}, pages
  2208--2217. PMLR, 2017.

\bibitem{lu2017unsupervised}
Hao Lu, Lei Zhang, Zhiguo Cao, Wei Wei, Ke Xian, Chunhua Shen, and Anton
  van~den Hengel.
\newblock When unsupervised domain adaptation meets tensor representations.
\newblock In {\em Proceedings of the IEEE International Conference on Computer
  Vision}, pages 599--608, 2017.

\bibitem{motiian2017unified}
Saeid Motiian, Marco Piccirilli, Donald~A Adjeroh, and Gianfranco Doretto.
\newblock Unified deep supervised domain adaptation and generalization.
\newblock In {\em Proceedings of the IEEE International Conference on Computer
  Vision}, pages 5715--5725, 2017.

\bibitem{pytorch}
Adam Paszke, Sam Gross, Soumith Chintala, Gregory Chanan, Edward Yang, Zachary
  DeVito, Zeming Lin, Alban Desmaison, Luca Antiga, and Adam Lerer.
\newblock Automatic differentiation in pytorch.
\newblock In {\em NeurIPS Workshop}, 2017.

\bibitem{multipeng2019moment}
Xingchao Peng, Qinxun Bai, Xide Xia, Zijun Huang, Kate Saenko, and Bo Wang.
\newblock Moment matching for multi-source domain adaptation.
\newblock In {\em Proceedings of the IEEE/CVF International Conference on
  Computer Vision}, pages 1406--1415, 2019.

\bibitem{redmon2016you}
Joseph Redmon, Santosh Divvala, Ross Girshick, and Ali Farhadi.
\newblock You only look once: Unified, real-time object detection.
\newblock In {\em Proceedings of the IEEE conference on computer vision and
  pattern recognition}, pages 779--788, 2016.

\bibitem{redmon2017yolo9000}
Joseph Redmon and Ali Farhadi.
\newblock Yolo9000: better, faster, stronger.
\newblock In {\em Proceedings of the IEEE conference on computer vision and
  pattern recognition}, pages 7263--7271, 2017.

\bibitem{ren2015faster}
Shaoqing Ren, Kaiming He, Ross Girshick, and Jian Sun.
\newblock Faster r-cnn: Towards real-time object detection with region proposal
  networks.
\newblock In {\em Advances in neural information processing systems}, pages
  91--99, 2015.

\bibitem{strong-weak}
Kuniaki Saito, Yoshitaka Ushiku, Tatsuya Harada, and Kate Saenko.
\newblock Strong-weak distribution alignment for adaptive object detection.
\newblock In {\em {IEEE} Conference on Computer Vision and Pattern
  Recognition}, 2019.

\bibitem{foggy}
Christos Sakaridis, Dengxin Dai, and Luc~Van Gool.
\newblock Semantic foggy scene understanding with synthetic data.
\newblock {\em International Journal of Computer Vision}, 126(9):973--992,
  2018.

\bibitem{simonyan2014very}
Karen Simonyan and Andrew Zisserman.
\newblock Very deep convolutional networks for large-scale image recognition.
\newblock {\em arXiv preprint arXiv:1409.1556}, 2014.

\bibitem{tzeng2015simultaneous}
Eric Tzeng, Judy Hoffman, Trevor Darrell, and Kate Saenko.
\newblock Simultaneous deep transfer across domains and tasks.
\newblock In {\em Proceedings of the IEEE International Conference on Computer
  Vision}, pages 4068--4076, 2015.

\bibitem{tzeng2017adversarial}
Eric Tzeng, Judy Hoffman, Kate Saenko, and Trevor Darrell.
\newblock Adversarial discriminative domain adaptation.
\newblock In {\em Proceedings of the IEEE conference on computer vision and
  pattern recognition}, pages 7167--7176, 2017.

\bibitem{van2008visualizing}
Laurens Van~der Maaten and Geoffrey Hinton.
\newblock Visualizing data using t-sne.
\newblock {\em Journal of machine learning research}, 9(11), 2008.

\bibitem{viola2001rapid}
Paul Viola and Michael Jones.
\newblock Rapid object detection using a boosted cascade of simple features.
\newblock In {\em Proceedings of the 2001 IEEE computer society conference on
  computer vision and pattern recognition. CVPR 2001}, volume~1, pages I--I.
  IEEE, 2001.

\bibitem{survey_2018}
Mei Wang and Weihong Deng.
\newblock Deep visual domain adaptation: A survey.
\newblock {\em Neurocomputing}, 312:135--153, 2018.

\bibitem{xu_cvpr20_icr_ccr}
Chang-Dong Xu, Xing-Ran Zhao, Xin Jin, and Xiu-Shen Wei.
\newblock Exploring categorical regularization for domain adaptive object
  detection.
\newblock In {\em Proceedings of the IEEE/CVF Conference on Computer Vision and
  Pattern Recognition}, pages 11724--11733, 2020.

\bibitem{GPA}
Minghao Xu, Hang Wang, Bingbing Ni, Qi Tian, and Wenjun Zhang.
\newblock Cross-domain detection via graph-induced prototype alignment.
\newblock In {\em Proceedings of the IEEE/CVF Conference on Computer Vision and
  Pattern Recognition}, pages 12355--12364, 2020.

\bibitem{multixu2018deep}
Ruijia Xu, Ziliang Chen, Wangmeng Zuo, Junjie Yan, and Liang Lin.
\newblock Deep cocktail network: Multi-source unsupervised domain adaptation
  with category shift.
\newblock In {\em Proceedings of the IEEE Conference on Computer Vision and
  Pattern Recognition}, pages 3964--3973, 2018.

\bibitem{multizhao2018adversarial}
Han Zhao, Shanghang Zhang, Guanhang Wu, Jos{\'e}~MF Moura, Joao~P Costeira, and
  Geoffrey~J Gordon.
\newblock Adversarial multiple source domain adaptation.
\newblock {\em Advances in neural information processing systems},
  31:8559--8570, 2018.

\bibitem{multizhao2019}
Sicheng Zhao, Bo Li, Xiangyu Yue, Yang Gu, Pengfei Xu, Runbo Hu, Hua Chai, and
  Kurt Keutzer.
\newblock Multi-source domain adaptation for semantic segmentation.
\newblock In {\em Advances in Neural Information Processing Systems}, pages
  7285--7298, 2019.

\bibitem{zheng_cvpr20_prototype}
Yangtao Zheng, Di Huang, Songtao Liu, and Yunhong Wang.
\newblock Cross-domain object detection through coarse-to-fine feature
  adaptation.
\newblock In {\em Proceedings of the IEEE/CVF Conference on Computer Vision and
  Pattern Recognition}, pages 13766--13775, 2020.

\bibitem{zhu_cvpr19_selective_alignment}
Xinge Zhu, Jiangmiao Pang, Ceyuan Yang, Jianping Shi, and Dahua Lin.
\newblock Adapting object detectors via selective cross-domain alignment.
\newblock In {\em {IEEE} Conference on Computer Vision and Pattern
  Recognition}, 2019.

\bibitem{zhuang_2020_ifan}
Chenfan Zhuang, Xintong Han, Weilin Huang, and M. Scott.
\newblock ifan: Image-instance full alignment networks for adaptive object
  detection.
\newblock In {\em AAAI}, 2020.

\end{thebibliography}
